\documentclass[conference]{IEEEtran}
\pdfoutput=1
\usepackage{times}

\usepackage{algorithmicx}
\usepackage{algorithm}
\usepackage{algpseudocode}

\usepackage{amsmath}
\usepackage{amssymb}
\usepackage{amsfonts}
\usepackage{amsbsy}
\usepackage{mathrsfs}
\usepackage{mathtools}

\newtheorem{proposition}{Proposition}

\usepackage{booktabs}
\usepackage{array}
\usepackage{makecell}

\usepackage{multirow}

\DeclareMathOperator*{\argmax}{\arg\!\max} 

\newcommand*\diff{\mathop{}\!\mathrm{d}}

\usepackage[numbers]{natbib}
\usepackage{multicol}
\usepackage[bookmarks=true]{hyperref}

\usepackage{xcolor}

\renewcommand{\baselinestretch}{0.98}

\begin{document}

\title{\huge Deep Visual Reasoning: Learning to Predict Action Sequences for Task and Motion Planning from an Initial Scene Image}

\author{
	\authorblockN{
		Danny Driess \hspace{2cm}
		Jung-Su Ha \hspace{2cm}
		Marc Toussaint
	}
	\vspace{0.2cm}
	\authorblockA{	
		Machine Learning and Robotics Lab, University of Stuttgart, Germany
	}
	\authorblockA{
		Max-Planck Institute for Intelligent Systems, Stuttgart, Germany
	}
	\authorblockA{	
		Learning and Intelligent Systems Group, TU Berlin, Germany
	}
	\vspace{-1.05cm}
}

\maketitle

\begin{abstract}
In this paper, we propose a deep convolutional recurrent neural network that predicts action sequences for task and motion planning (TAMP) from an initial scene image. Typical TAMP problems are formalized by combining reasoning on a symbolic, discrete level (e.g.\ first-order logic) with continuous motion planning such as nonlinear trajectory optimization.
Due to the great combinatorial complexity of possible discrete action sequences, a large number of optimization/motion planning problems have to be solved to find a solution, which limits the scalability of these approaches.

To circumvent this combinatorial complexity, we develop a neural network which, based on an initial image of the scene, directly predicts promising discrete action sequences such that ideally only one motion planning problem has to be solved to find a solution to the overall TAMP problem.
A key aspect is that our method generalizes to scenes with many and varying number of objects, although being trained on only two objects at a time.
This is possible by encoding the objects of the scene in images as input to the neural network, instead of a fixed feature vector.
Results show runtime improvements of several magnitudes.
Video: \url{https://youtu.be/i8yyEbbvoEk}

\end{abstract}

\IEEEpeerreviewmaketitle
\section{Introduction}\label{sec:intro}
\vspace{-0.1cm}
A major challenge of sequential manipulation problems is that they inherently involve discrete and continuous aspects. 
To account for this hybrid nature of manipulation, Task and Motion Planning (TAMP) problems are usually formalized by combining reasoning on a symbolic, discrete level with continuous motion planning.
The symbolic level, e.g.\ defined in terms of first-order logic, proposes high level discrete action sequences for which the motion planner, for example nonlinear trajectory optimization or a sampling-based method, tries to find motions that fulfill the requirements induced by the high level action sequence or return that the sequence is infeasible.

Due to the high combinatorial complexity of possible discrete action sequences, a large number of motion planning problems have to be solved to find a solution to the TAMP problem.
This is mainly caused by the fact that many TAMP problems are difficult, since the majority of action sequences are actually infeasible, mostly due to kinematic limits or geometric constraints.
Moreover, it takes more computation time for a motion planner to reliably detect infeasibility of a high level action sequence than to find a feasible motion when it exists.
Consequently, sequential manipulation problems, which intuitively seem simple, can take a very long time to solve.

To overcome this combinatorial complexity, we aim to learn to predict promising action sequences from the scene as input. 
Using this prediction as a heuristic on the symbolic level, we can
drastically reduce the number of motion planning problems that need to be evaluated.
Ideally, we seek to directly predict a \emph{feasible} action sequence, requiring only a \emph{single} motion planning problem to be solved.

However, learning to predict such action sequences imposes multiple challenges.
First of all, the objects in the scene and the goal have to be encoded as input to the predictor in a way that enables similar generalization capabilities to classical TAMP approaches with respect to scenes with many and changing number of objects and goals.
Secondly, the large variety of such scenes and goals, especially if multiple objects are involved, makes it difficult to generate a sufficient dataset.

Recently, \cite{wells2019learning} and \cite{driess20deep} propose a classifier that predicts the feasibility of a motion planning problem resulting from a discrete decision in the task domain.
However, a major limitation of their approaches is that the feasibility for only a \emph{single} action is predicted, whereas the combinatorial complexity of TAMP especially arises from action \emph{sequences} and it is not straightforward to utilize such a classifier for action sequence prediction within TAMP.

\begin{figure}
	\includegraphics[]{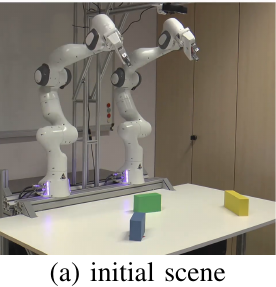}
	\includegraphics[]{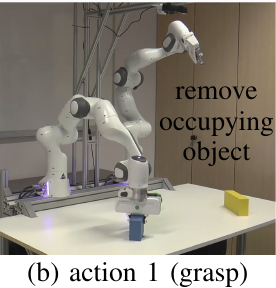}
	\includegraphics[]{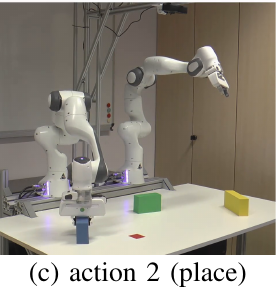}
	\includegraphics[]{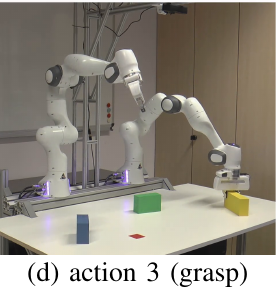}
	\includegraphics[]{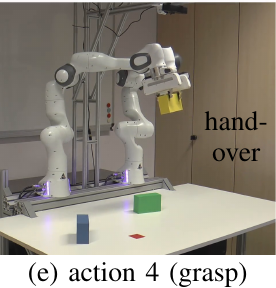}
	\includegraphics[]{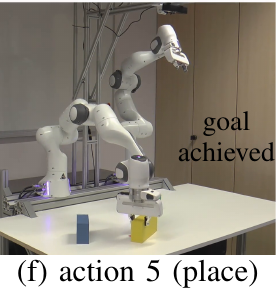}
	\vspace{-0.3cm}
	\caption{Typical scene: The yellow object should be placed on the red spot, which is, however, occupied by the blue object. Furthermore, the yellow object cannot be reached by the robot arm that is able to place it on the red spot.}
	\label{fig:firstPage}
	\vspace{-0.56cm}
\end{figure}

To address these issues, we develop a neural network that predicts action \emph{sequences} from the initial scene \emph{and} the goal as input.
An important question is how the objects in the scene can be encoded as input to the predictor in a way that shows similar generalization capabilities of classical TAMP approaches.
By encoding the objects (and the goal) in the image space, we show that the network is able to generalize to scenes with many and changing number of objects with only little runtime increase, although it has been trained on only a fixed number of objects.
Compared to a purely discriminative model, since the predictions of our network are goal-conditioned, we do not need to use the network to search over many sequences, but can directly generate them with it.

The predicted action sequences parameterize a nonlinear trajectory optimization problem that optimizes a globally consistent paths fulfilling the requirements induced by the actions.

\noindent To summarize, our main contributions are
\begin{itemize}
	\item A convolutional, recurrent neural network that predicts from an initial scene image and a task goal promising action sequences, which parameterize a nonlinear trajectory optimization problem, to solve the TAMP problem.
	\item A way to integrate this network into the tree search algorithm of the underlying TAMP framework.
	\item We demonstrate that the network generalizes to situations with many and varying numbers of objects in the scene, although it has been trained on only two objects at a time.
\end{itemize}
From a methodological point of view, this work contains nonlinear trajectory optimization, first-order logic reasoning and deep convolutional, recurrent neural networks.

\section{Related Work}\label{sec:relatedWork}
\vspace{-0.09cm}
\subsection{Learning to Plan}
\vspace{-0.05cm}
There is great interest in learning to mimic planning itself.
The architectures in \citep{tamar2016value,okada2017path,srinivas18universal,amos2018differentiable} resemble value iteration, path integral control, gradient-based trajectory optimization and iterative LQR methods, respectively.
For sampling-based motion planning, \citep{ichter2018learning} learn an optimal sampling distribution conditioned on the scene and the goal to speed up planning.
To enable planning with raw sensory input, there are several works that learn a compact representation and its dynamics in sensor space to then apply planning or reinforcement learning (RL) in the learned latent space~\citep{boots11closing,xieimprovisation, ha2018adaptive,ichter2019robot, watter15embed, finn16deep, lange12autonomous, silver2017predictron}.
Another line of research is to learn an action-conditioned predictive model \cite{finn17deep,xieimprovisation,paxton19visual, ebert17self, dosovitskiy17learning, pascanu17learning, racaniere2017imagination}.
With this model, the future state of the environment for example in image space conditioned on the action is predicted, which can then be utilized within MPC \citep{finn17deep,xieimprovisation} or to guide tree search \citep{paxton19visual}.
The underlying idea is that learning the latent representation and dynamics enables reasoning with high-dimensional sensory data.
However, a disadvantage of such predictive models is that still a search over actions is necessary, which grows exponentially with sequence length. 
For our problem which contains handovers or other complex behaviors that are induced by an action, learning a predictive model in the image space seems difficult. 
Most of these approaches focus on low level actions.
Furthermore, the behavior of our trajectory optimizer is only defined for a complete action sequence, since future actions have an influence on the trajectory of the past.
Therefore, state predictive models cannot directly be applied to our problem.

The proposed method in the present work learns a relevant representation of the scene from an initial scene image such that a recurrent module can reason about long-term action effects without a direct state prediction.

\subsection{Learning Heuristics for TAMP and MIP in Robotics}
A general approach to TAMP is to combine discrete logic search with a sampling-based motion planning algorithm~\citep{kaelbling2010hierarchical, silva13towards, srivastava14combined, dantam18ijrr} or constraint satisfaction methods \cite{lagriffoul12constraint, lagriffoul14efficiently, lozanoPerez14constraint}.
A major difficulty arises from the fact that the number of feasible symbolic sequences increases exponentially with the number of objects and sequence length.
To reduce the large number of geometric problems that need to be solved, many heuristics have been developed, e.g.\ \cite{kaelbling2010hierarchical, rodriguez19iteratively, 19-driess-RSSws}, to efficiently prune the search tree.
Another approach to TAMP is Logic Geometric Programming (LGP) \citep{toussaint15lgp,toussaint17mbts,18-toussaint-RSS, 20-toussaint-physicsLGP, 20-ha-PLGP}, which combines logic search with trajectory optimization.
The advantage of an optimization based approach to TAMP is that the trajectories can be optimized with global consistency, which, e.g., allows to generate handover motions efficiently.
LGP will be the underlying framework of the present work.
For large-scale problems, however, LGP also suffers from the exponentially increasing number of possible symbolic action sequences \cite{hartmann2020robust}.
Solving this issue is one of the main motivations for our work.

Instead of handcrafted heuristics, there are several approaches to integrate learning into TAMP to guide the discrete search in order to speed up finding a solution \cite{garret, chitnis16guided, kim2019learning, kim2018guiding, wang2018active}.
However, these mainly act as heuristics, meaning that one still has to search over the discrete variables and probably solve many motion planning problems.
In contrast, the network in our approach generates goal-conditioned action sequences, such that in most cases there is no search necessary at all.
Similarly, in optimal control for hybrid domains mixed-integer programs suffer from the same combinatorial complexity \citep{hogan2016feedback,hogan18reactive,doshi2020hybrid}.
LGP also can be viewed as a generalization of mixed-integer programs.
In \cite{carpentier17learning} (footstep planning) and \cite{hogan18reactive} (planar pushing), learning is used to predict the integer assignments, however, this is for a single task only with no generalization to different scenarios.

A crucial question in integrating learning into TAMP is how the scene and goals can be encoded as input to the learning algorithm in a way that enables similar generalization capabilities of classical TAMP.
For example, in \citep{paxton19visual} the considered scene contains always the same four objects with the same colors, which allows them to have a fixed input vector of separate actions for all objects.
In \cite{wilson2019collections} convolutional (CNN) and graph neural networks are utilized to learn a state representation for RL, similarly in \cite{li2020towards}.
In \cite{bejjani2019learning}, rendered images from a simulator are used as state representation to exploit the generalization ability of CNNs.
In our work, the network learns a representation in image space
that is able to reason over complex action sequences from an initial observation only and is able to generalize over changing numbers of objects.

The work of \citet{wells2019learning} and \citet{driess20deep} is most related to our approach.
They both propose to learn a classifier which predicts the feasibility of a motion planning problem resulting from a \emph{single} action. 
The input is a feature representation of the scene \cite{wells2019learning} or a scene image \cite{driess20deep}.
While both show generalization capabilities to multiple objects, one major challenge of TAMP comes from action \emph{sequences} and it is, however, unclear how a single step classifier as in \cite{wells2019learning} and \cite{driess20deep} could be utilized for sequence prediction.

To our knowledge, the our work is the first that learns to generate action sequences for an optimization based TAMP approach from an initial scene image and the goal as input, while showing generalization capabilities to multiple objects.

\section{Logic Geometric Programming for \\ Task and Motion Planning}\label{sec:LGP}
This work relies on Logic Geometric Programming (LGP) \cite{toussaint15lgp, toussaint17mbts} as the underlying TAMP framework.
The main idea behind LGP is a nonlinear trajectory optimization problem over the continuous variable $x$, in which the constraints and costs are parameterized by a discrete variable $s$ that represents the state of a symbolic domain.
The transitions of this variable are subject to a first-order logic language that induces a decision tree.
Solving an LGP involves a tree search over the discrete variable, where each node represents a nonlinear trajectory optimization program (NLP).
If a symbolic leaf node, i.e.\ a node which state $s$ is in a symbolic goal state, is found and its corresponding NLP is feasible, a solution to the TAMP problem has been obtained.
In this section, we briefly describe LGP for the purpose of this work.

Let $\mathcal{X} = \mathcal{X}(S)\subset \mathbb{R}^{n(S)}\times SE(3)^{m(S)}$ be the configuration space of all objects and articulated structures (robots) as a function of the scene $S$. 
The idea is to find a global path $x$ in the configuration space which minimizes the LGP
\begin{subequations}\label{eq:LGP}
	\begin{align}
	&P(g, S) =
	\!\!\min_{ \substack{K\in\mathbb{N} \\x:[0,KT]\rightarrow \mathcal{X} \\ a_{1:K},~s_{1:K} } }  \int_{0}^{KT} \!\!\!\!\!c\big(x(t), \dot{x}(t), \ddot{x}(t), s_{k(t)}, S\big) \diff t  \label{eq:LGP:objective} 
	\\[-0.3cm]
	&~~~~\text{ s.t.}\notag
	\end{align}
	\vspace{-0.9cm}
	\begin{align}
	&\forall_{t\in[0, KT]} : && h_\text{eq}\big(x(t), \dot{x}(t), s_{k(t)}, S\big) = 0 \label{eq:LGP:eq}\\
	&\forall_{t\in[0, KT]} : && h_\text{ineq}\big(x(t), \dot{x}(t), s_{k(t)}, S\big) \le 0 \label{eq:LGP:ineq}\\
	&\forall_{k=1,\ldots,K} : && h_\text{sw}\big(x(kT), \dot{x}(kT), a_k, S\big) = 0\label{eq:LGP:switch}\\
	&\forall_{k=1,\ldots,K} : && a_k \in \mathbb{A}(s_{k-1}, S)\label{eq:LGP:A}\\
	&\forall_{k=1,\ldots,K} : && s_k = \text{succ}(s_{k-1}, a_k)\label{eq:LGP:succ}\\
	&&& x(0) = \tilde{x}_0(S)\label{eq:LGP:init}\\
	&&& s_0 = \tilde{s}_0(S)\label{eq:LGP:initX}\\
	&&& s_K \in \mathcal{S}_\mathrm{goal}(g). \label{eq:LGP:goal}
	\end{align}
\end{subequations}
The path is assumed to be globally continuous ($x\in C([0, TK], \mathcal{X})$) and consists of $K\in\mathbb{N}$ phases (the number is part of the decision problem itself), each of fixed duration $T>0$, in which we require smoothness $x\in C^2([(k-1)T, kT])$.
These phases are also referred to as kinematic modes \cite{1985-Mason-mechanicsmanipulation,18-toussaint-RSS}.
The functions $c,h_\text{eq},h_\text{ineq}$ and hence the objectives in phase $k$ of the motion ($k(t) = \lfloor t/T \rfloor$) are parameterized by the discrete variable (or integers in mixed-integer programming) $s_k\in\mathcal{S}$, representing the state of the symbolic domain.
The time discrete transitions between $s_{k-1}$ and $s_k$ are determined by the successor function $\text{succ}(\cdot, \cdot)$, which is a function of the previous state $s_{k-1}$ and the discrete action $a_k$ at phase $k$.
The actions are grounded action operators.
Which actions are possible at which symbolic state is determined by the logic and expressed in the set $\mathbb{A}(s_{k-1}, S)$.
$h_\text{sw}$ is a function that imposes transition conditions between the kinematic modes.
The task or goal of the TAMP problem is defined symbolically through the set $\mathcal{S}_\text{goal}(g)$ for the symbolic goal (a set of grounded literals) $g\in\mathbb{G}(S)$, e.g.\ placing an object on a table.
The quantities $\tilde{x}_0(S)$ and $\tilde{s}_0(S)$ are the scene dependent initial continuous and symbolic states, respectively.
For fixed $s$ it is assumed that $c$, $h_\text{eq}$ and $h_\text{ineq}$ are differentiable.
Finally, we define the feasibility of an action sequence $a_{1:K}=(a_1,\ldots, a_K)$ as 
\begin{align}
	F_S\left(a_{1:K}\right) = \begin{cases}
	 1 & \exists x:[0,KT]\rightarrow \mathcal{X} :\eqref{eq:LGP:eq}-\eqref{eq:LGP:initX}\\
	 0 & \text{else}
	\end{cases}
\end{align}

\subsection{Multi-Bound LGP Tree Search and Lower Bounds}
The logic induces a decision tree (called LGP-tree) through \eqref{eq:LGP:A} and \eqref{eq:LGP:succ}.
Solving a path problem as a heuristic to guide the tree search is too expensive.
A key contribution of \cite{toussaint17mbts} is therefore to introduce relaxations or lower bounds on \eqref{eq:LGP} in the sense that the feasibility of a lower bound is a necessary condition on the feasibility of the complete problem \eqref{eq:LGP}, while these lower bounds should be computationally faster to compute. Each node in the LGP tree defines several lower bounds of \eqref{eq:LGP}. 
Still, as we will show in the experiments, a large number of NLPs have to be solved to find a feasible solution for problems with a high combinatorial complexity.
This is especially true if many decisions are feasible in early phases of the sequence, but then later become infeasible.

\section{Deep Visual Reasoning}\label{sec:DGR}
The main idea of this work is, given the scene and the task goal as input, to predict a promising discrete action sequence $a_{1:K} = (a_1, \ldots, a_K)$ which reaches a symbolic goal state and its corresponding trajectory optimization problem is feasible.
An ideal algorithm would directly predict an action sequence such that only a single NLP has to be solved to find an overall feasible solution, which consequently would lead to a significant speedup in solving the LGP \eqref{eq:LGP}.

We will first describe more precisely what should be predicted, then how the scene, i.e.\ the objects and actions with them, and the goal can be encoded as input to a neural network that should perform the prediction.
Finally, we discuss how the network is integrated into the tree search algorithm in a way that either directly predicts a feasible sequence or, in case the network is mistaken, acts as a heuristic to further guide the search without losing the ability to find a solution if one exists.
We additionally propose an alternative way to integrate learning into LGP based on a recurrent feasibility classifier.

\subsection{Predicting Action Sequences}
First of all, we define for the goal $g$ the set of all action sequences that lead to a symbolic goal state in the scene $S$ as
\begin{align}
	\mathcal{T}\left(g, S\right)\! =\! \big\{ a_{1:K} :~ & \forall_{i=1}^K~ a_i\in \mathbb{A}(s_{i-1}, S), ~s_{i} = \mathrm{succ}(s_{i-1}, a_i)\notag\\
	& s_0 = \tilde{s}_0(S), ~ s_K \in \mathcal{S}_\mathrm{goal}(g)\big\}.	
\end{align}
In relation to the LGP-tree, this is the set of all leaf nodes and hence candidates for an overall feasible solution.
One idea is to learn a discriminative model which predicts whether a complete sequence leads to a feasible NLP and hence to a solution.
To predict an action sequence one would then choose the sequence from $\mathcal{T}(g, S)$ where the discriminative model has the highest prediction.
However, computing $\mathcal{T}(\cdot, \cdot)$ (up to a maximum $K$) and then checking all sequences with the discriminative model is computationally inefficient.

Instead, we propose to learn a function $\pi$ (a neural network) that, given a scene description $S$, the task goal $g$ and the past decisions $a_{1:k-1}$, predicts whether an action $a_k$ at the current time step $k$ is promising in the sense of the probability that there exist future actions $a_{k+1:K}$ such that the complete sequence $a_{1:K}$ leads to a feasible NLP that solves the original TAMP problem.
Formally,
\begin{align}
	&\pi\big(a_k,~ g, ~a_1, \ldots, a_{k-1},~ S \big) = \notag \\
	&~~~~~p\Big(\exists_{K\ge k} \exists_{a_{k+1}, \ldots, a_K } ~:~ a_{1:K} \in \mathcal{T}\left(g, S\right), ~F_S\left(a_{1:K} \right) = 1 \notag \\ 
	& ~~~~~~~~~~~~~~~~~~~~~~~~~~~~~~~~~\Big| ~a_k, ~g, ~a_1, \ldots, a_{k-1}, ~S\Big).\!\label{eq:pi}
\end{align} 
This way, $\pi$ generates an action sequence by choosing the action at each step where $\pi$ has the highest prediction.

\begin{figure} 
	\centering
	\scalebox{0.9}{
		\includegraphics[]{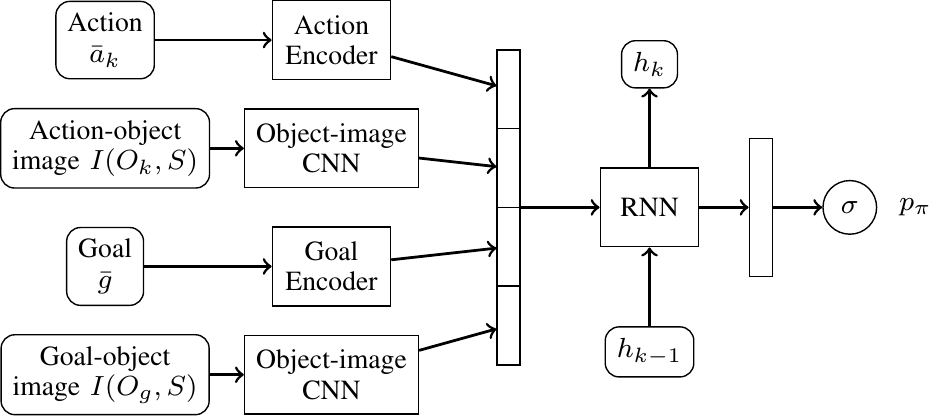}
	}
	\vspace{-0.7cm}
	\caption{Proposed neural network architecture.}
	\vspace{-0.6cm}
	\label{fig:networkArchitecture}
\end{figure}

\subsection{Training Targets}\label{sec:DGR:trainingTargets}
The crucial question arises how $\pi$ as defined in \eqref{eq:pi} can be trained.
The semantics of $\pi$ is related to a universal Q-function, but it evaluates actions $a_k$ based on an implicit representation of state (see Sec.~\ref{sec:QFunction}). Furthermore, it turns out that we can cast the problem into supervised learning by transforming the data into suitable training targets.
Assume that one samples scenes $S^i$, goals $g^i$ as well as goal-reaching action sequences $a_{1:K^i}^i \in \mathcal{T}\left(g^i, S^i\right)$, e.g.\ with breadth-first search. For each of these sampled sequences, the feasibility of the resulting NLP is determined and saved in the set
\begin{align}
\mathcal{D}_\mathrm{data}& = \Big\{ \Big(S^i,a_{1:K^i}^i, g^i, F_{S^i}\left(a_{1:K^i}^i\right) \Big) \Big\}_{i=1}^n.
\end{align}
Based on this dataset, we define the \emph{training} dataset for $\pi$ as
\begin{align}
	\mathcal{D}_\mathrm{train} = \Big\{ \Big(S^i, a_{1:K^i}^i, g^i, f^i \Big) \Big\}_{i=1}^n
\end{align}
where $f^i\in\left\{0,1\right\}^{K_i}$ is a sequence of binary labels. Its $j$th component $f^i_j$ indicates for every \emph{sub}sequence $a_{1:j}^i$ whether it should be classified as promising as follows
\begin{align}
	f^i_j =
	\begin{cases}
		1 & F_{S^i}\left( a_{1:K^i}^i \right) = 1 \\
		1 & \exists~{ \left( S^l, a_{1:K^l}^l, g^l, F^l \right) \in \mathcal{D}_\mathrm{data} } : \\
		& ~~~~~~~~~~\text{ }~F^l = F_{S^l}\left( a_{1:K^l}^l \right) = 1 \\
		& ~~~~~~~~\wedge ~ g^l = g^i ~\wedge ~ a_{1:j}^l = a_{1:j}^i\\
		0 & \text{else}
	\end{cases}\label{eq:trainingTargets}
\end{align}
If the action sequence is feasible and solves the problem specified by
$g^i$, then $f_j^i = 1$ for all $j=1,\ldots,K_i$ (first case).  This
is the case where $\pi$ should predict a high probability at each step
$k$ to follow a feasible sequence.  If the action sequence with index
$i$ is \emph{not} feasible, but there exists a feasible one in
$\mathcal{D}_\mathrm{data}$ (index $l$) which has an overlap with the
other sequence up to step $j$, i.e.\ $a_{1:j}^l = a_{1:j}^i$, then
$f_j^i = 1$ as well (second case).  Also in this case the network
should suggest to follow this decision, since it predicts that there
exist future decisions which lead to a feasible solution.  Finally, in
the last and third case where the sequence is infeasible and has no
overlap with other feasible sequences, $f_j^i = 0$, meaning that the
network should predict to not follow this decision.  This data
transformation is a simple pre-processing step that allows us to train
$\pi$ in a supervised sequence labeling setting, with the standard
(weighted) binary cross-entropy loss. Another advantageous
side-effect of this transformation is that it creates a more balanced
dataset with respect to the training targets.

\subsection{Input to the Neural Network -- Encoding $a$, $g$ and $S$}
So far, we have formulated the predictor $\pi$ in \eqref{eq:pi} in terms of the scene $S$, symbolic actions $a$ and the goal $g$ of the LGP \eqref{eq:LGP}.
In order to represent $\pi$ as a neural network, we need to find a suitable encoding of $a$, $g$ and $S$.

\subsubsection{Splitting actions into action operator symbols and objects}
An action $a$ is a grounded action operator, i.e.\ it is a combination of an action operator symbol and objects in the scene it refers to, similarly for a goal $g$.
While the number of action operators is assumed to be constant, the number of objects can be completely different from scene to scene.
Most neural networks, however, expect inputs of fixed dimension.
In order to achieve the same generalization capabilities of TAMP approaches with respect to changing numbers of objects, we encode action and goal symbols very differently to the objects they operate on.
In particular, object references are encoded in a way that includes geometric scene information.

Specifically, given an action $a$, we decompose it into $a=(\bar{a}, O)\in\mathcal{AO}(s, S)\subset\mathcal{A}\times\mathcal{P}(\mathcal{O}(S))$, where $\bar{a}\in\mathcal{A}$ is its discrete action operator symbol and $O\in\mathcal{P}(\mathcal{O}(S))$ the \emph{tuple} of objects the action operates on.
The goal is similarly decomposed into $g = (\bar{g}, O_g)$, $\bar{g}\in\mathcal{G}$, $O_g\in\mathcal{P}(\mathcal{O}(S))$.
This separation seems to be a minor technical detail, which is, however, of key importance for the generalizability of our approach to scenes with changing numbers of objects.

Through that separation, since, as mentioned before, the cardinality of $\mathcal{A}$ and $\mathcal{G}$ is constant and independent from the scene, we can input $\bar{a}$ and $\bar{g}$ directly as a one-hot encoding to the neural network.

\subsubsection{Encoding the objects $O$ and $O_g$ in the image space}
For our approach it is crucial to encode the information about the objects in the scene in a way that allows the neural network to generalize over objects (number of objects in the scene and their properties).
By using the separation of the last paragraph, we can introduce the mapping $I: (O, S) \mapsto \mathbb{R}^{(n_c + n_{O}) \times w \times h}$, which encodes any scene $S$ and object tuple $O$ to a so-called action-object-image encoding, namely an $n_c + n_O$-channel image of width $w$ and height $h$, where the first $n_c$ channels represent an image of the \emph{initial} scene and the last $n_{O}$ channels are binary masks which indicate the subset of objects that are involved in the action.
These last mask channels not only encode object identity, but substantial geometric and relational information, which is a key for the predictor to predict feasible action sequences.
In the experiments, the scene image is a depth image, i.e.\ $n_c = 1$ and the maximum number of objects that are involved in a single action is two, hence $n_O = 2$.
If an action takes less objects into account than $n_O$, this channel is zeroed. 
Since the maximum number $n_O$ depends on the set of actions operator symbols $\mathcal{A}$, which has a fixed cardinality independent from the scene, this is no limitation.
The masks create an attention mechanism which is the key to generalize to multiple objects \cite{driess20deep}.
However, since each action object image $I(O, S)$ always contains a channel providing information of the complete scene, also the geometric relations to other objects are taken into account.
Being able to generate such masks is a reasonable assumption, since there are many methods for that.
Please note that these action-object-images always correspond to the \emph{initial} scene.

\subsubsection{Network Architecture}
Fig.~\ref{fig:networkArchitecture} shows the network architecture that represents $\pi$ as a convolutional recurrent neural network.
Assume that in step $k$ the probability should be predicted whether an action $a_k = (\bar{a}_k, O_k)$ for the goal $g = (\bar{g}, O_g)$ in the scene $S$ is promising.
The action object images $I(O_k, S)$ as well as the goal object images $I(O_g, S)$ are encoded by a convolutional neural network (CNN).
The discrete action/goal symbols $\bar{a}, \bar{g}$ are encoded by fully connected layers with a one-hot encoding as input.
Since the only information the network has access to is the \emph{initial} configuration of the scene, a recurrent neural network (RNN) takes the current encoding of step $k$ and the past encodings, which it has to learn to represent in its hidden state $h_{k-1}$, into account.
Therefore, the network has to implicitly generate its own predictive model about the effects of the actions, without explicitly being trained to reproduce some future state.
The symbolic goal $\bar{g}$ and its corresponding goal object image $I(O_g, S)$ are fed into the neural network at each step, since it is constant for the complete task.
The weights of the CNN action-object-image encoder can be shared with the CNN of the goal-object-image encoder, since they operate on the same set of object-images.
To summarize,
\begin{align}
	(p_\pi, h_k) &=\pi_\text{NN}\big(\bar{a}_k, I(O_k,S), \bar{g}, I(O_g,S), h_{k-1}\big) \notag \\ & =\pi\big(a_k, g, a_{1:k-1}, S\big)
\end{align}

\subsection{Relation to Q-Functions}\label{sec:QFunction}
In principle, one can view the way we define $\pi$ in \eqref{eq:pi} and how we propose to train it with the transformation \eqref{eq:trainingTargets} as learning a goal-conditioned Q-function in a partially observable Markov decision process (POMDP), where a binary reward of 1 is assigned if a complete action sequence is feasible and reaches the symbolic goal.
However, there are important differences.
For example, a Q-function usually relies on a clear notion of state.
In our case, the symbolic state $s$ does not contain sufficient information, since it neither includes geometry nor represents the effects of all past decisions on the NLP. Similarly for the continuous state $x$, which is only defined for a complete action sequence.  
Therefore, our network has to learn a state representation from the past action-object image sequence, while only observing the initial state in form of the depth image of the scene as input.
Furthermore, we frame learning $\pi$ as a supervised learning problem.

\subsection{Algorithm}
Algo.~\ref{algo:LGPNN} presents the pseudocode how $\pi$ is integrated in the tree search algorithm.
The main idea is to maintain the set $E$ of expand nodes.
A node $n = \left(s,(\bar{a},O), k, p_\pi, h, n_\text{parent}\right)$ in the tree consists of its symbolic state $s(n)$, action-object pair $(\bar{a},O)$, depth $k(n)$, i.e.\ the current sequence length, the prediction of the neural network $p_\pi(n)$, the hidden state of the neural network $h(n)$ and the parent node $n_\text{parent}$.
At each iteration, the algorithm chooses the node $n^\ast_E$ of the expand list where the network has the highest prediction (line \ref{algo:LGPNN:expandHighest}).
For all possible next actions, i.e.\ children of $n^\ast_E$, the network is queried to predict their probability leading to a feasible solution, which creates new nodes (line \ref{algo:LGPNN:newNode}).
If a node reaches a symbolic goal state (line \ref{algo:LGPNN:goalState}), it is added to the set of leaf nodes $L$, else to the expand set.
Then the already found leaf nodes are investigated.
Since during the expansion of the tree leaf nodes which are unlikely to be feasible are also found, only those trajectory optimization problems are solved where the prediction $p_\pi$ is higher than the feasibility threshold $f_\text{thresh}$ (set to 0.5 in the experiments).
This reduces the number of NLPs that have to be solved.
However, one cannot expect that the network never erroneously has a low prediction although the node would be feasible.
In order to prevent not finding a feasible solution in such cases, the function $\text{adjustFeasibilityThreshold}(\cdot)$ reduces this threshold with a discounting factor or sets it to zero if all leaf nodes with a maximum depth of $K_\text{max}$ have been found.
This gives us the following.
\begin{proposition}
	Algorithm 1 is complete in the sense that if a scene contains at least one action sequence with maximum length $K_\text{max}$ for which the nonlinear trajectory optimizer can find a feasible motion, the neural network does not prevent finding this solution, even in case of prediction errors.
\end{proposition}

As an important remark, we store the hidden state of the recurrent neural network in its corresponding node.
Furthermore, the object image and action encodings also have to be computed only once.
Therefore, during the tree search, only one pass of the recurrent (and smaller) part of the complete $\pi_\text{NN}$ has to be queried in each step.

\begin{algorithm}[t]
	\caption{LGP with Deep Visual Reasoning}
	\label{fig:algo}
	\begin{algorithmic}[1]
		\State\textbf{Input:} Scene $S$, goal $g$ and max sequence length $K_\text{max}$
		\State $L = \emptyset$ \Comment{{\footnotesize set of leaf nodes}}
		\State $E = \{n_0\}$ \Comment{{\footnotesize set of nodes to be expanded, $n_0$ is root node}}
		\While {no solution found}
			\Statex \hskip\algorithmicindent {\footnotesize $\triangleright$ choose node from expand set with highest prediction}
			\State $\displaystyle n_E^\ast = \argmax_{ n\in E ~\wedge~ k(n) < K_\text{max} } p_\pi(n) $ \label{algo:LGPNN:expandHighest}
			\State $E \leftarrow E \backslash \{n_E^\ast\}$
			\ForAll {$(\bar{a}, O) \in \mathcal{AO}(s(n_E^\ast), S)$}
				\State $(p_\pi, h) = \pi_\text{NN}\left(\bar{a}, I(O,S), \bar{g}, I(O_g, S), h(n^\ast_E) \right)$
				\State $s = \mathrm{succ}(s(n_E^\ast), (\bar{a},O)), ~~k = k(n_E^\ast) + 1$
				\State $n = \left(s,(\bar{a},O), k, p_\pi, h, n^\ast_E\right)$\label{algo:LGPNN:newNode} \Comment{{\footnotesize new node}}
				\If {$s \in \mathcal{S}_\text{goal}(g)$ }\label{algo:LGPNN:goalState}
					\State $L \leftarrow L \cup \{n\}$ \Comment{{\footnotesize if goal state, add to leaf node set}}
				\Else
					\State $E\rightarrow E\cup \left\{n\right\}$ \Comment{{\footnotesize if no goal state, add to expand set}}
				\EndIf
			\EndFor
			\While {$|L| > 0$}  \Comment{{\footnotesize consider already found leaf nodes}}
				\Statex \hskip\algorithmicindent \hskip\algorithmicindent {\footnotesize $\triangleright$ choose node from leaf node set with highest prediction}
				\State $\displaystyle n_L^\ast = \argmax_{n\in L}~p_\pi(n)$
				\If {$p_\pi(n_L^\ast) \le f_\text{thresh}$ }
					\State $f_\text{thresh}\leftarrow\text{adjustFeasibilityThreshold}(f_\text{thresh})$
					\State \textbf{break}
				\EndIf
				\State $L \leftarrow L \backslash \{n^\ast_L\}$
				\State $(\bar{a},O)_{1:K} = (\bar{a},O)_{1:k(n^\ast_L)}(n^\ast_L)$\Comment{{\footnotesize extract action seq.}}
				\State solve NLP $x = P\left( (\bar{a},O)_{1:K}, g, O_g, S\right)$
				\If {feasible, i.e.\ $F_S((\bar{a}, O)_{1:K}) = 1$}
					\State solution $(\bar{a},O)_{1:K}$ with trajectory $x$ found
					\State\textbf{break}
				\EndIf
			\EndWhile
		\EndWhile
	\end{algorithmic}
	\label{algo:LGPNN}
\end{algorithm}

\subsection{Alternative: Recurrent Feasibility Classifier}
The method of \cite{driess20deep} and \cite{wells2019learning} to learn a feasibility classifier considers single actions only, i.e.\ no sequences.
To allow for a comparision we here present an approach to extend the idea of a feasibility classifier to action sequences and how it can be integrated into our TAMP framework.
The main idea is to classify the feasibility of an action sequence with a recurrent classifier, independently from the fact whether it has reached a symbolic goal or not.
This way, during the tree search solving an NLP can be replaced by evaluating the classifier, which usually is magnitudes faster.
Technically, this classifier $\pi_\text{RC}\left(\bar{a}_k, I(O_k,S), h_{k-1}\right)$ has a similar architecture as $\pi_\text{NN}$, but only takes the current action-object-image pair as well as the hidden state of the previous step as input and predicts whether the action sequence up to this step is feasible.
A disadvantage is that just because an action is feasible does not necessarily mean that following it will solve the TAMP problem in the long term.
Sec.~\ref{sec:exp:compToClassifier} presents an empirical comparison.

\section{Experiments}\label{sec:exp}
The video \url{https://youtu.be/i8yyEbbvoEk} demonstrates the planned motions both in simulation and with a real robot.

\begin{figure}
	\centering
	\includegraphics[trim={19cm, 14cm, 12cm, 16cm}, clip, width=2.cm]{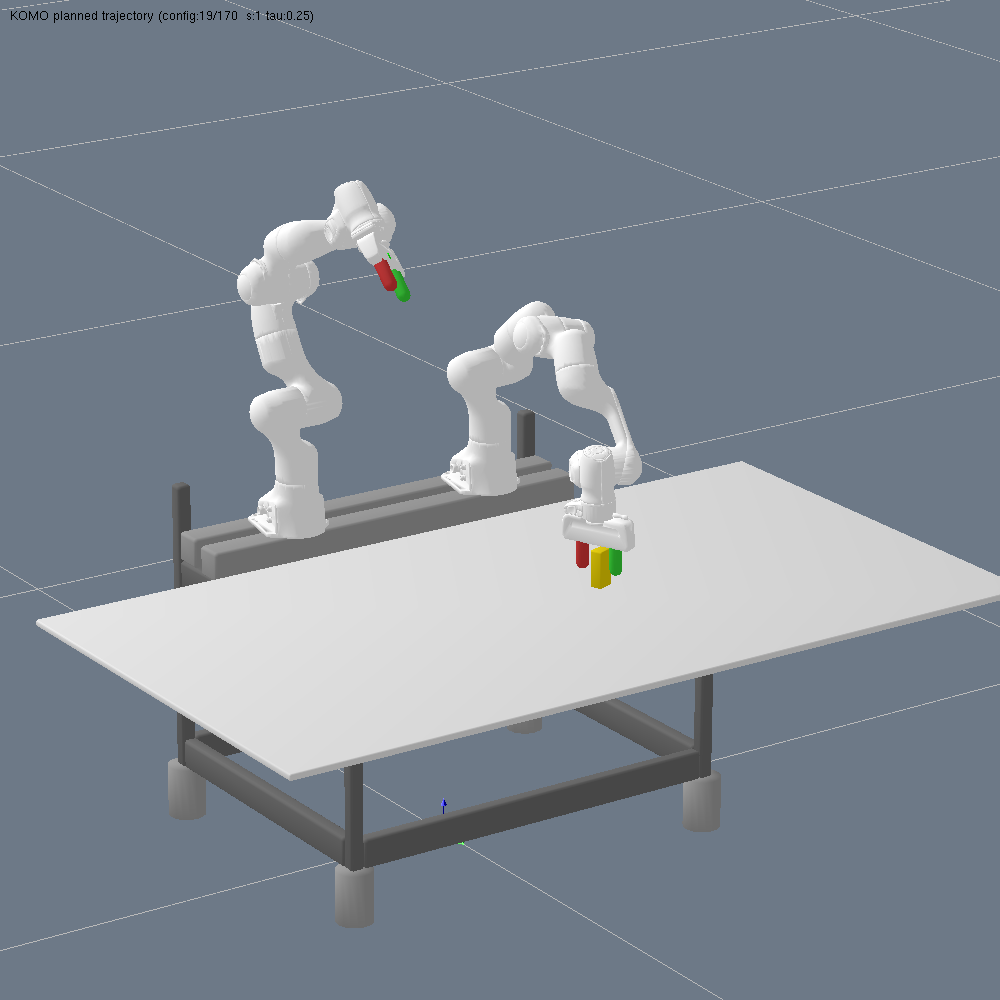}
	\includegraphics[trim={19cm, 14cm, 12cm, 16cm}, clip, width=2.cm]{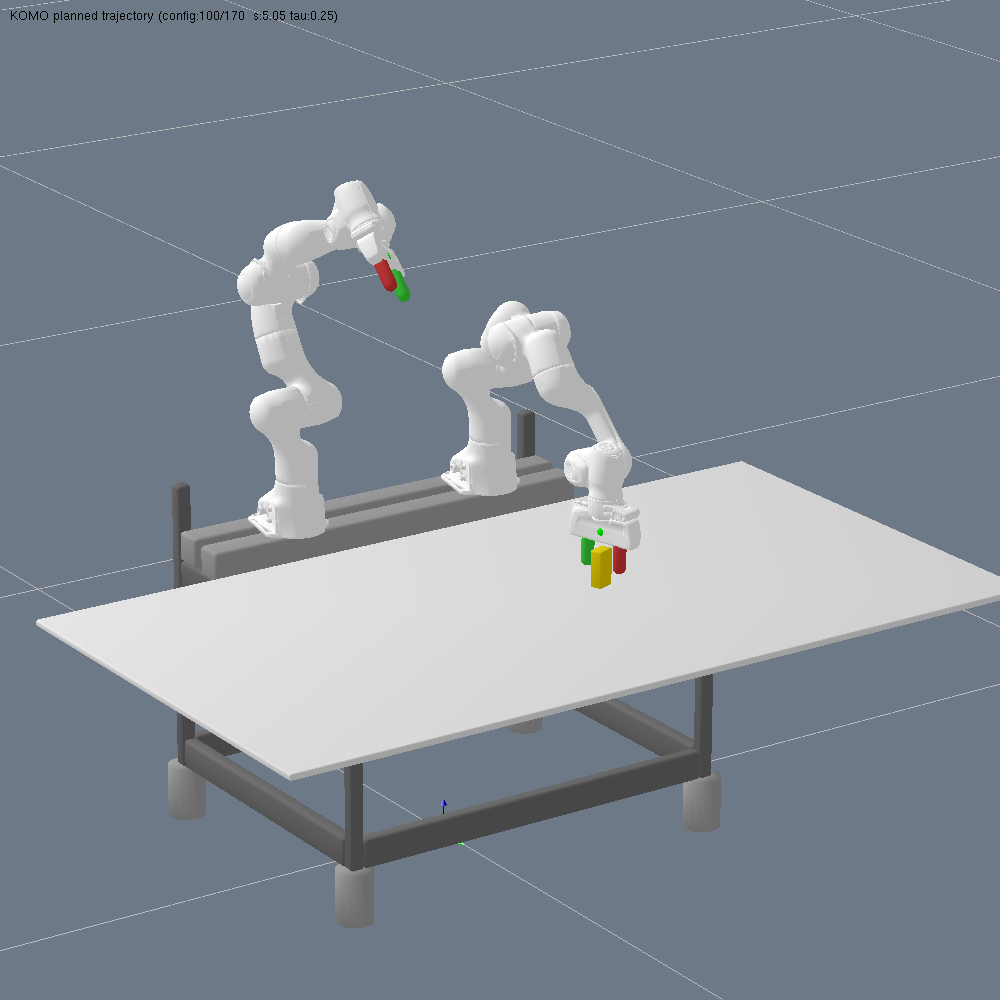}
	\includegraphics[trim={19cm, 14cm, 12cm, 16cm}, clip, width=2.cm]{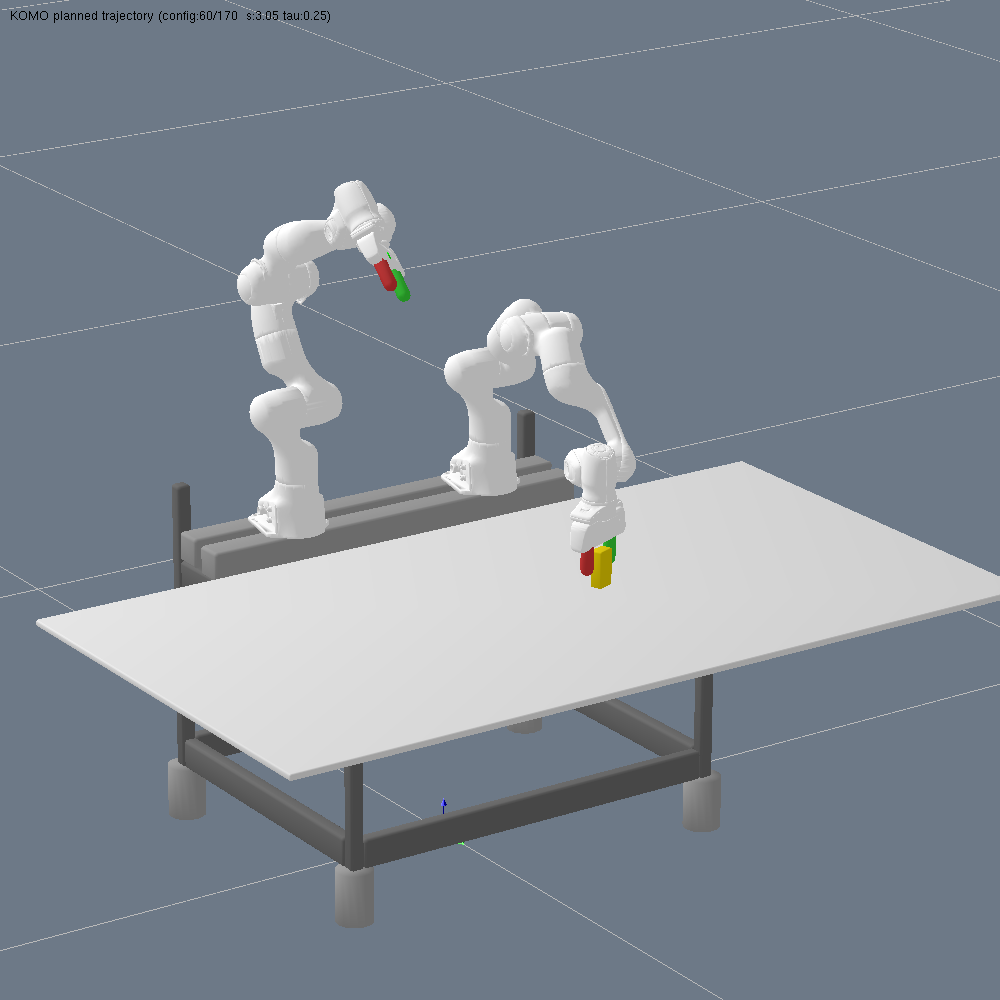}
	\includegraphics[trim={19cm, 14cm, 12cm, 16cm}, clip, width=2.cm]{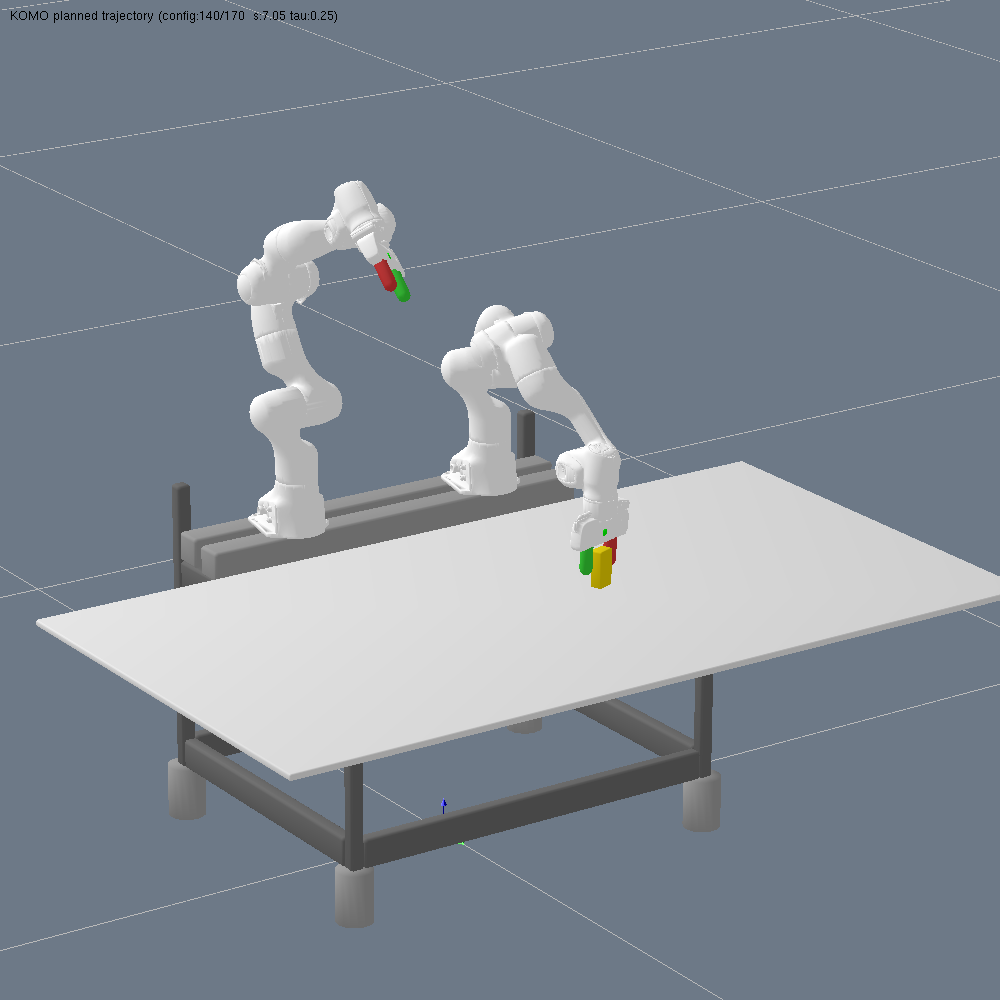}
	\vspace{-0.3cm}
	\caption{The four different integer assignments of the \texttt{grasp} operator.}
	\vspace{-0.6cm}
	\label{fig:exp:graspings}
\end{figure}

\subsection{Setup and Task}
We consider a tabletop scenario with two robot arms (Franka Emika Panda) and multiple box-shaped objects, see Fig.~\ref{fig:firstPage} for a typical scene, in which the goal is to move an object to different target locations, visualized by a red square in Fig.~\ref{fig:firstPage}.

\subsubsection{Action Operators and Optimization Objectives}
The logic is described by PDDL-like rules.
There are two action operators \texttt{grasp} and \texttt{place}.
The \texttt{grasp} action takes as parameters the robot arm and one of four integers $\eta$, represented in its discrete action symbol $\bar{a}$, and one object $O$.
Depending on the integer, the end-effector is aligned to different surfaces of the box (equality constraint).
Furthermore, an inequality constraint ensures that the center point between the two grippers is inside of the object (with a margin).
In Fig.~\ref{fig:exp:graspings} these four discrete ways of grasping are visualized for one robot arm.
The exact grasping location relative to the object is still subject to the optimizer.
The \texttt{place} action has the robot arm (also encoded in the discrete symbol $\bar{a}$) and two objects as the tuple $O$ as parameters.
The effects on the optimization objectives are that the bottom surface of object one touches and is parallel to object two.
In our case, object one is a box, whereas object two is the table or the goal location.
Preconditions for \texttt{grasp} and \texttt{place} ensure that one robot arm attempts to grasp only one object simultaneously and that an arm can only place an object if it is holding one.
Path costs are squared accelerations on $x$.
There are collisions and joint limits as inequality constraints (with no margin).

\subsubsection{Properties of the Scene}
There are multiple properties which make this (intuitively simple) task challenging for task and motion planning algorithms.
First of all, the target location can fully or partially be occupied by another object.
Secondly, the object and/or the target location can be out of reach for one of the robot arms. 
Hence, the algorithm has to figure out which robot arm to use at which phase of the plan and the two robot arms possibly have to collaborate to solve the task.
Thirdly, apart from position and orientation, the objects vary in size, which also influences the ability to reach or place an object.
In addition, grasping box-shaped objects introduces a combinatorics that is not handled well by nonlinear trajectory optimization due to local minima and also joint limits.
Therefore, as described in the last paragraph, we introduce integers as part of the discrete action that influence the grasping geometry.
This greatly increases the branching factor of the task.
For example, depending on the size of the object, it has to be grasped differently or a handover between the two arms is possible or not, which has a significant influence on the feasibility of action sequences.

Indeed, Tab.~\ref{tab:numberOfLeafNodesOverDepth} shows the number of action sequences with a certain length that lead to a symbolic goal state over the number of objects in the scene.
This number corresponds to \emph{candidate} sequences for a feasible solution (the set $\mathcal{T}\left(g, S\right)$) which demonstrates the great combinatorial complexity of the task, not only with respect to sequence length, but also number of objects.
One could argue that an occupied and reachability predicate could be introduced  in the logic to reduce the branching of the tree.
However, this requires a reasoning engine which decides those predicates for a given scene, which is not trivial for general cases.
More importantly, both reachability and occupation by another object is something that is also dependent on the geometry of the object that should be grasped or placed and hence not something that can be precomputed in all cases \cite{driess20deep, 19-driess-IROS}.
For example, if the object that is occupying the target location is small and  the object that should be placed there also, then it can be placed directly, while a larger object that should be placed requires to first remove the occupying object.
Our algorithm does not rely on such non-general simplifications, but decides promising action sequences based on the real relational geometry of the scene.

\begin{table}
	\caption{Number of action sequences that reach a symbolic goal state}
	\vspace{-0.3cm}
	\centering
	\begin{tabular}{crrrrr}
		\toprule
		\multicolumn{1}{c|}{\# of objects} & \multicolumn{5}{c}{length of the action sequence} \\
		\multicolumn{1}{c|}{in the scene} & \multicolumn{1}{c}{\hspace{0.3cm}2} & \multicolumn{1}{c}{3} & \multicolumn{1}{c}{4} & \multicolumn{1}{c}{5} & \multicolumn{1}{c}{6} \\
		\midrule
		1 & 8 & 32 & 192 & 1,024 & 5,632\\
		2 & 8 & 96 & 704 & 6,400 & 51,200 \\
		3 & 8 & 160 & 1,216 & 15,872 & 145,920\\
		4 & 8 & 224 & 1,728 & 29,440 & 289,792\\
		5 & 8 & 288 & 2,240 & 47,104 & 482,816\\
		\bottomrule
	\end{tabular}
	\label{tab:numberOfLeafNodesOverDepth}
	\vspace{-0.5cm}
\end{table}

\subsection{Training/Test Data Generation and Network Details}\label{sec:exp:TrainingData}
We generated 30,000 scenes with two objects present at a time.
The sizes, positions and orientations of the objects as well as the target location are sampled uniformly within a certain range.
For half of the scenes, one of the objects (not the one that is part of the goal) is placed directly on the target, to ensure that at least half of the scenes contain a situation where the target location is occupied.
The dataset $\mathcal{D}_\text{data}$ is determined by a breadth-first search for each scene over the action sequences, until either 4 solutions have been found or 1,000 leaf nodes have been considered.
In total, for 25,736 scenes at least one solution was found, which were then the scenes chosen to create the actual training dataset $\mathcal{D}_\text{train}$ as described in Sec.~\ref{sec:DGR:trainingTargets}.
102,566 of the action sequences in $\mathcal{D}_\text{data}$ were feasible, 2,741,573 completely infeasible.
This shows the claim of the introduction that the majority of action sequences are actually infeasible.
Furthermore, such an imbalanced dataset imposes difficulties for a learning algorithm.
With the data transformation from Sec.~\ref{sec:DGR:trainingTargets}, there are 7,926,696 $f_j = 0$ and 1,803,684 $f_j = 1$ training targets in $\mathcal{D}_\text{train}$, which is more balanced.

The network is trained with the ADAM optimizer (learning rate 0.0005) with a batch size of 48.
To account for the aforementioned imbalance in the dataset, we oversample feasible action sequences such that at least 16 out of the 48 samples in one batch come from a feasible sequence.
The image encoder consists of three convolutional layers with 5, 10, 10 channels, respectively, and filter size of 5x5.
The second and third convolutional layer has a stride of 2.
After the convoultional layers, there is a fully connected layer with an output feature size of 100.
The same image encoder is used to encode the action images and the goal image.
The discrete action encoder is one fully connected layer with 100 neurons and relu activations.
The recurrent part consists of one layer with 300 GRU cells, followed by a linear layer with output size 1 and a sigmoid activation as output for $\pi$.
Since the task is always to place an object at varying locations, we left out the discrete goal encoder in the experiments presented here.

To evaluate the performance and accuracy of our method, we sampled 3000 scenes, again containing two objects each, with the same algorithm as for the training data, but with a different random seed.
Using breadth-first search, we determined 2705 feasible scenes, which serve as the actual test scenes.

\subsection{Performance -- Results on Test Scenarios}
\begin{figure}
	\centering
	\includegraphics[]{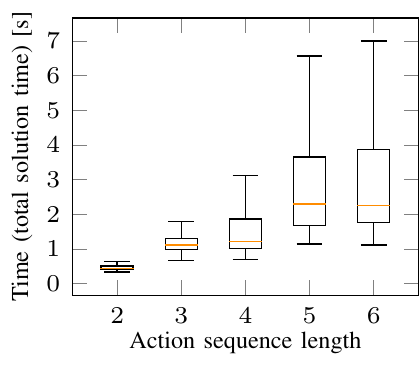}
	\includegraphics[]{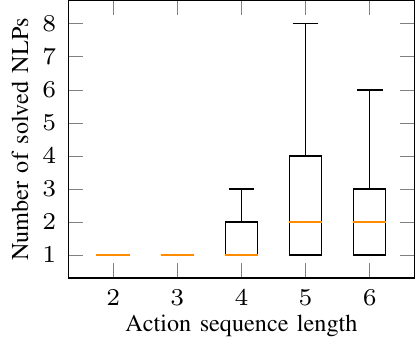}
	\vspace{-0.4cm}
	\caption{Total time (left) and number of solved NLPs (right) to find an overall feasible solution for the test scenes with neural network}
	\label{fig:exp:performanceNN}
	\vspace{-0.3cm}
\end{figure}

Fig.~\ref{fig:exp:performanceNN} shows both the total runtime and the number of NLPs that have to be solved to find a feasible solution. 
When we report the total runtime, we refer to everything, meaning computing the image/action encodings, querying the neural network during the search and all involved NLPs that are solved.
As one can see, for all cases with sequence lengths of 2 and 3, the first predicted sequence is feasible, such that there is no search and only one NLP has to be solved.
For length 3, the median is still 1, but also for sequences of lengths 5 and 6 in half of the cases less than two NLPs have to be solved.
Generally, with a median runtime of about 2.3 s for even sequence length of 6, the overall framework with the neural network has a high performance.
Furthermore, the upper whiskers are also below 7 s.
All experiments have been performed with an Intel Xeon W-2145 CPU @ 3.70GHz.

\subsection{Comparison to Multi-Bound LGP Tree Search}
\begin{figure}
	\centering
	\includegraphics[]{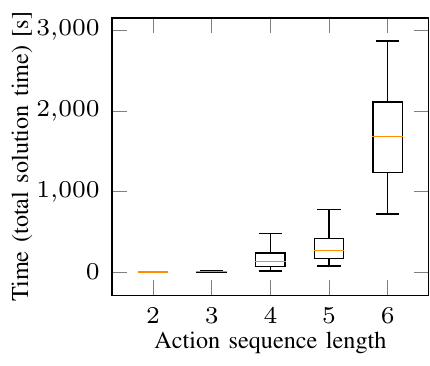}
	\includegraphics[]{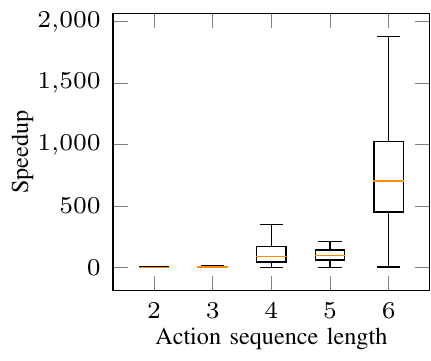}%
	\vspace{-0.4cm}
	\caption{Comparison to LGP tree search. Left: total runtime. Right: the speedup of our neural network (sol.\ time with NN / sol.\ time with LGP tree search).}
	\label{fig:exp:compLGP}
	\vspace{-0.5cm}
\end{figure}

In Fig.~\ref{fig:exp:compLGP} (left) the runtimes for solving the test cases with LGP tree search are presented, which shows the difficulty of the task.
In 132 out of the 2,705 test cases, LGP tree search is not able to find a solution within the timeout, compared to only 3 with the neural network.
Fig.~\ref{fig:exp:compLGP} (right) shows the speedup that is gained by using the neural network.
For sequence length 4, the network is 46 times faster, 100 times for length 5 and for length 6 even 705 times (median).
In this plot, only those scenes where LGP tree search and the neural network have found the same sequence lengths are compared.

\subsection{Comparison to Recurrent Classifier}\label{sec:exp:compToClassifier}
\begin{figure}
	\centering
	\includegraphics[]{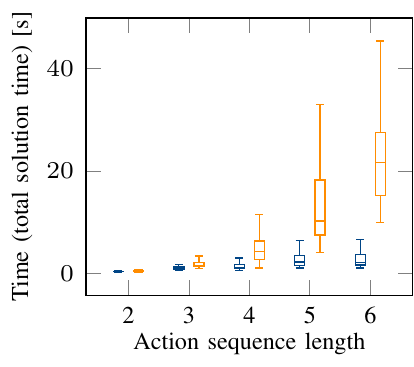}
	\includegraphics[]{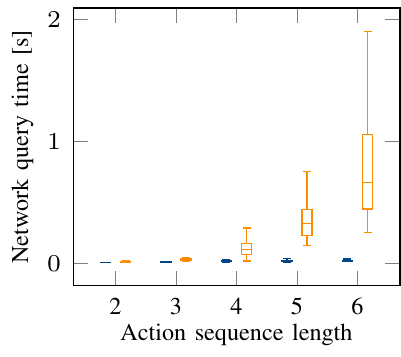}
	\vspace{-0.4cm}
	\caption{Comparison to recurrent classifier (orange). Blue is our network.}
	\label{fig:exp:RC}
	\vspace{-0.6cm}
\end{figure}

Fig.~\ref{fig:exp:RC} (left) shows a comparison of our proposed goal-conditioned network that generates sequences to a recurrent classifier that only predicts feasibility of an action sequence, independent from the task goal.
As one can see, while such a classifier also leads to a significant speedup compared to LGP tree search, our goal-conditioned network has an even higher speedup, which also stays relatively constant with respect to increasing action sequence lengths. 
Furthermore, with the classifier 22 solutions have not been found, compared to 3 with our approach.
While the network query time is neglectable for our network, as can be seen in Fig.~\ref{fig:exp:RC} (right), the time to query the recurrent classifier becomes visible.

\subsection{Generalization to Multiple Objects}\label{sec:exp:multipleObjects}
Creating a rich enough dataset containing combinations of different numbers of objects is infeasible.
Instead, we now take the network that has been trained as described in Sec.~\ref{sec:exp:TrainingData} with only two objects present at a time and test whether it generalizes to scenes with more than two (and also only one) objects.
The 200 test scenes are always the same, but more and more objects are added.
Fig.~\ref{fig:exp:generalization}a reports the total runtime to find a feasible solution with our proposed neural network over the number of objects present in the scene.
These runtimes include all scenes with different action sequence lengths.
There was not a single scene where no solution is found.
While the upper quartile increases, the median is not significantly affected by the presence of multiple objects.
Generally, the performance is remarkable, especially when looking at Tab.~\ref{tab:numberOfLeafNodesOverDepth} where for sequence length 6 there are nearly half a million possible action sequences.
The fact that the runtime increases for more objects is not only caused by the fact that the network inevitable does some mistakes and hence more NLPs have to be solved.
Solving (even a feasible) NLP with more objects can take more time due to increased collision queries and increased non-convexity.

\subsection{Generalization to Cylinders}
\begin{figure}
	\centering
	\includegraphics[]{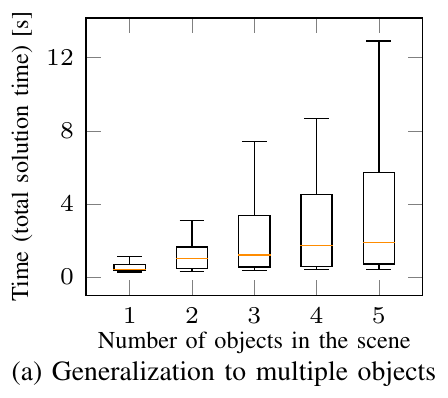}
	\hspace{-0.4cm}
	\includegraphics[]{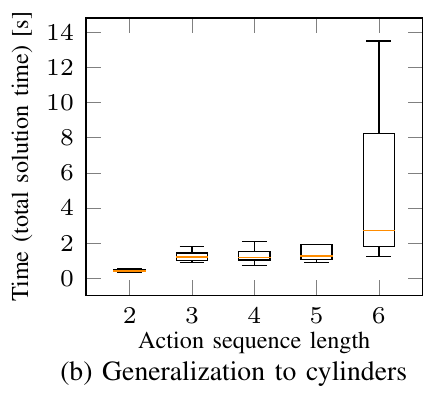}
	\vspace{-0.4cm}
	\caption{Generalization experiments.}
	\label{fig:exp:generalization}
	\vspace{-0.6cm}
\end{figure}

Although the network has been trained on box-shaped objects only, we investigate if the network can generalize to scenes which contain other shapes like cylinders.
Since the objects are encoded in the image space, there is a chance that, as compared to a feature space which depends on a less general parameterization of the shape, this is possible.
We generated 200 test scenes that either contain two cylinders, three cylinders or a mixture of a box and a cylinder, all of different sizes/positions/orientations and targets.
If the goal is to place a cylinder on the target, we made sure in the data generation that the cylinder has an upper limit on its radius in order to be graspable.
These cylinders, however, have a relatively similar appearance in the rasterized image as boxes.  
Therefore, the scenes also contain cylinders which have larger radii such that they have a clearly different appearance than what is contained in the dataset.
Fig.~\ref{fig:exp:generalization}b shows the total solution time with the neural network.
As one can see, except for action sequences of length 6, there is no drop in performance compared to box-shaped objects, which indicates that the network is able to generalize to other shapes.
Even for length 6, the runtimes are still very low, especially compared to LGP tree search. 
Please note that our constraints for the nonlinear trajectory optimization problem are general enough to deal with boxes and cylinders.
However, one also has to state that for even more general shapes the trajectory optimization becomes a problem in its own.

\subsection{Real Robot Experiments}\label{sec:exp:realRobot}
Fig.~\ref{fig:firstPage} shows our complete framework in the real world. 
In this scene the blue object occupies the goal location and the target object (yellow) is out of reach for the robot arm that is be able to place it on the goal.
Since the yellow object is large enough, the network proposed a handover solution (Fig.~\ref{fig:firstPage}c).
The presence of an additional object (green) does not confuse the predictions.
The planned trajectories are executed open-loop.
The images as input to the neural network are rendered from object models obtained by a perception pipeline, therefore, the transfer to the real robot is directly possible. 

\section{Conclusion}\label{sec:conclusion}
In this work, we proposed a neural network that learned to predict promising discrete action sequences for TAMP from an initial scene image and the task goal as input.
In most cases, the first sequence generated by the network was feasible.
Hence, despite the fact that the network could act as a heuristic, there was no real search over the discrete variable and consequently only one trajectory optimization problem had to be solved to find a solution to the TMAP problem.

Although being trained on only two objects present at a time, the learned representation of the network enabled to generalize to scenes with multiple objects (and other shapes to some extend) while still showing a high performance.

The main assumption and therefore limitation of the proposed method is that the initial scene image has to contain sufficient information to solve the task, which means no total occlusions or other ambiguities.

\section*{Acknowledgments}
Danny Driess thanks the International Max-Planck Research School for Intelligent Systems (IMPRS-IS) for the support. Marc Toussaint thanks the Max-Planck Fellowship at MPI-IS.

\cleardoublepage

\bibliographystyle{plainnat}
\bibliography{references}

\begin{thebibliography}{50}
\providecommand{\natexlab}[1]{#1}
\providecommand{\url}[1]{\texttt{#1}}
\expandafter\ifx\csname urlstyle\endcsname\relax
  \providecommand{\doi}[1]{doi: #1}\else
  \providecommand{\doi}{doi: \begingroup \urlstyle{rm}\Url}\fi

\bibitem[Amos et~al.(2018)Amos, Jimenez, Sacks, Boots, and
  Kolter]{amos2018differentiable}
Brandon Amos, Ivan Jimenez, Jacob Sacks, Byron Boots, and J~Zico Kolter.
\newblock Differentiable mpc for end-to-end planning and control.
\newblock In \emph{Advances in Neural Information Processing Systems}, pages
  8289--8300, 2018.

\bibitem[Bejjani et~al.(2019)Bejjani, Dogar, and Leonetti]{bejjani2019learning}
W~Bejjani, MR~Dogar, and M~Leonetti.
\newblock Learning physics-based manipulation in clutter: Combining image-based
  generalization and look-ahead planning.
\newblock In \emph{International Conference on Intelligent Robots and Systems
  (IROS)}. IEEE, 2019.

\bibitem[Boots et~al.(2011)Boots, Siddiqi, and Gordon]{boots11closing}
Byron Boots, Sajid~M Siddiqi, and Geoffrey~J Gordon.
\newblock Closing the learning-planning loop with predictive state
  representations.
\newblock \emph{The International Journal of Robotics Research}, 2011.

\bibitem[Carpentier et~al.(2017)Carpentier, Budhiraja, and
  Mansard]{carpentier17learning}
Justin Carpentier, Rohan Budhiraja, and Nicolas Mansard.
\newblock Learning feasibility constraints for multicontact locomotion of
  legged robots.
\newblock In \emph{Robotics: Science and Systems}, 2017.

\bibitem[Chitnis et~al.(2016)Chitnis, Hadfield{-}Menell, Gupta, Srivastava,
  Groshev, Lin, and Abbeel]{chitnis16guided}
Rohan Chitnis, Dylan Hadfield{-}Menell, Abhishek Gupta, Siddharth Srivastava,
  Edward Groshev, Christopher Lin, and Pieter Abbeel.
\newblock Guided search for task and motion plans using learned heuristics.
\newblock In \emph{International Conference on Robotics and Automation (ICRA)},
  pages 447--454. {IEEE}, 2016.

\bibitem[Dantam et~al.(2018)Dantam, Kingston, Chaudhuri, and
  Kavraki]{dantam18ijrr}
Neil~T. Dantam, Zachary~K. Kingston, Swarat Chaudhuri, and Lydia~E. Kavraki.
\newblock An incremental constraint-based framework for task and motion
  planning.
\newblock \emph{International Journal on Robotics Research}, 2018.

\bibitem[de~Silva et~al.(2013)de~Silva, Pandey, Gharbi, and
  Alami]{silva13towards}
Lavindra de~Silva, Amit~Kumar Pandey, Mamoun Gharbi, and Rachid Alami.
\newblock Towards combining {HTN} planning and geometric task planning.
\newblock \emph{CoRR}, 2013.

\bibitem[Doshi et~al.(2020)Doshi, Hogan, and Rodriguez]{doshi2020hybrid}
Neel Doshi, Francois~R Hogan, and Alberto Rodriguez.
\newblock Hybrid differential dynamic programming for planar manipulation
  primitive.
\newblock In \emph{International Conference on Robotics and Automation (ICRA)}.
  {IEEE}, 2020.

\bibitem[Dosovitskiy and Koltun(2017)]{dosovitskiy17learning}
Alexey Dosovitskiy and Vladlen Koltun.
\newblock Learning to act by predicting the future.
\newblock In \emph{International Conference on Learning Representations
  {ICLR}}, 2017.

\bibitem[Driess et~al.(2019{\natexlab{a}})Driess, Oguz, and
  Toussaint]{19-driess-RSSws}
Danny Driess, Ozgur Oguz, and Marc Toussaint.
\newblock Hierarchical task and motion planning using logic-geometric
  programming ({HLGP}).
\newblock RSS Workshop on Robust Task and Motion Planning, 2019{\natexlab{a}}.

\bibitem[Driess et~al.(2019{\natexlab{b}})Driess, Schmitt, and
  Toussaint]{19-driess-IROS}
Danny Driess, Syn Schmitt, and Marc Toussaint.
\newblock Active inverse model learning with error and reachable set estimates.
\newblock In \emph{Proc{.} of the IEEE International Conference on Intelligent
  Robots and Systems (IROS)}, 2019{\natexlab{b}}.

\bibitem[Driess et~al.(2020)Driess, Oguz, Ha, and Toussaint]{driess20deep}
Danny Driess, Ozgur Oguz, Jung-Su Ha, and Marc Toussaint.
\newblock Deep visual heuristics: Learning feasibility of mixed-integer
  programs for manipulation planning.
\newblock In \emph{Proc. of the IEEE International Conference on Robotics and
  Automation (ICRA)}, 2020.

\bibitem[Ebert et~al.(2017)Ebert, Finn, Lee, and Levine]{ebert17self}
Frederik Ebert, Chelsea Finn, Alex~X. Lee, and Sergey Levine.
\newblock Self-supervised visual planning with temporal skip connections.
\newblock In \emph{Conference on Robot Learning}, 2017.

\bibitem[Finn and Levine(2017)]{finn17deep}
Chelsea Finn and Sergey Levine.
\newblock Deep visual foresight for planning robot motion.
\newblock In \emph{International Conference on Robotics and Automation (ICRA)},
  pages 2786--2793. {IEEE}, 2017.

\bibitem[Finn et~al.(2016)Finn, Tan, Duan, Darrell, Levine, and
  Abbeel]{finn16deep}
Chelsea Finn, Xin~Yu Tan, Yan Duan, Trevor Darrell, Sergey Levine, and Pieter
  Abbeel.
\newblock Deep spatial autoencoders for visuomotor learning.
\newblock In \emph{International Conference on Robotics and Automation (ICRA)}.
  {IEEE}, 2016.

\bibitem[Garrett et~al.(2016)Garrett, Kaelbling, and Lozano-Perez]{garret}
Caelan Garrett, Leslie Kaelbling, and Tomas Lozano-Perez.
\newblock Learning to rank for synthesizing planning heuristics.
\newblock In \emph{Proc. of the Int. Joint Conf. on Artificial Intelligence
  (IJCAI)}, 2016.

\bibitem[Ha et~al.(2018)Ha, Park, Chae, Park, and Choi]{ha2018adaptive}
Jung-Su Ha, Young-Jin Park, Hyeok-Joo Chae, Soon-Seo Park, and Han-Lim Choi.
\newblock Adaptive path-integral autoencoders: Representation learning and
  planning for dynamical systems.
\newblock In \emph{Advances in Neural Information Processing Systems}, pages
  8927--8938, 2018.

\bibitem[Ha et~al.(2020)Ha, Driess, and Toussaint]{20-ha-PLGP}
Jung-Su Ha, Danny Driess, and Marc Toussaint.
\newblock Probabilistic framework for constrained manipulations and task and
  motion planning under uncertainty.
\newblock In \emph{Proc. of the IEEE International Conference on Robotics and
  Automation (ICRA)}, 2020.

\bibitem[Hartmann et~al.(2020)Hartmann, Oguz, Driess, Toussaint, and
  Menges]{hartmann2020robust}
Valentin~N Hartmann, Ozgur~S Oguz, Danny Driess, Marc Toussaint, and Achim
  Menges.
\newblock Robust task and motion planning for long-horizon architectural
  construction planning.
\newblock arXiv:2003.07754, 2020.

\bibitem[Hogan and Rodriguez(2016)]{hogan2016feedback}
Fran{\c{c}}ois~Robert Hogan and Alberto Rodriguez.
\newblock Feedback control of the pusher-slider system: A story of hybrid and
  underactuated contact dynamics.
\newblock In \emph{Proceedings of the Workshop on Algorithmic Foundation
  Robotics (WAFR)}, 2016.

\bibitem[Hogan et~al.(2018)Hogan, Grau, and Rodriguez]{hogan18reactive}
Francois~Robert Hogan, Eudald~Romo Grau, and Alberto Rodriguez.
\newblock Reactive planar manipulation with convex hybrid {MPC}.
\newblock In \emph{International Conference on Robotics and Automation (ICRA)},
  pages 247--253. {IEEE}, 2018.
\newblock URL \url{https://doi.org/10.1109/ICRA.2018.8461175}.

\bibitem[Ichter and Pavone(2019)]{ichter2019robot}
Brian Ichter and Marco Pavone.
\newblock Robot motion planning in learned latent spaces.
\newblock \emph{Robotics and Automation Letters}, 4\penalty0 (3):\penalty0
  2407--2414, 2019.

\bibitem[Ichter et~al.(2018)Ichter, Harrison, and Pavone]{ichter2018learning}
Brian Ichter, James Harrison, and Marco Pavone.
\newblock Learning sampling distributions for robot motion planning.
\newblock In \emph{International Conference on Robotics and Automation (ICRA)},
  pages 7087--7094. IEEE, 2018.

\bibitem[Kaelbling and Lozano-P{\'e}rez(2011)]{kaelbling2010hierarchical}
Leslie~Pack Kaelbling and Tom{\'a}s Lozano-P{\'e}rez.
\newblock Hierarchical planning in the now.
\newblock In \emph{Proc. of the IEEE International Conference on Robotics and
  Automation (ICRA)}, 2011.

\bibitem[Kim et~al.(2018)Kim, Kaelbling, and Lozano-P{\'e}rez]{kim2018guiding}
Beomjoon Kim, Leslie~Pack Kaelbling, and Tom{\'a}s Lozano-P{\'e}rez.
\newblock Guiding search in continuous state-action spaces by learning an
  action sampler from off-target search experience.
\newblock In \emph{Thirty-Second AAAI Conference on Artificial Intelligence},
  2018.

\bibitem[Kim et~al.(2019)Kim, Wang, Kaelbling, and
  Lozano-P{\'e}rez]{kim2019learning}
Beomjoon Kim, Zi~Wang, Leslie~Pack Kaelbling, and Tom{\'a}s Lozano-P{\'e}rez.
\newblock Learning to guide task and motion planning using score-space
  representation.
\newblock \emph{The International Journal of Robotics Research}, 38\penalty0
  (7):\penalty0 793--812, 2019.

\bibitem[{Lagriffoul} et~al.(2012){Lagriffoul}, {Dimitrov}, {Saffiotti}, and
  {Karlsson}]{lagriffoul12constraint}
F.~{Lagriffoul}, D.~{Dimitrov}, A.~{Saffiotti}, and L.~{Karlsson}.
\newblock Constraint propagation on interval bounds for dealing with geometric
  backtracking.
\newblock In \emph{International Conference on Intelligent Robots and Systems},
  2012.

\bibitem[Lagriffoul et~al.(2014)Lagriffoul, Dimitrov, Bidot, Saffiotti, and
  Karlsson]{lagriffoul14efficiently}
Fabien Lagriffoul, Dimitar Dimitrov, Julien Bidot, Alessandro Saffiotti, and
  Lars Karlsson.
\newblock Efficiently combining task and motion planning using geometric
  constraints.
\newblock \emph{International Journal on Robotics Research}, 2014.

\bibitem[Lange et~al.(2012)Lange, Riedmiller, and
  Voigtl{\"{a}}nder]{lange12autonomous}
Sascha Lange, Martin~A. Riedmiller, and Arne Voigtl{\"{a}}nder.
\newblock Autonomous reinforcement learning on raw visual input data in a real
  world application.
\newblock In \emph{{IJCNN}}, 2012.

\bibitem[Li et~al.(2020)Li, Jabri, Darrell, and Agrawal]{li2020towards}
Richard Li, Allan Jabri, Trevor Darrell, and Pulkit Agrawal.
\newblock Towards practical multi-object manipulation using relational
  reinforcement learning.
\newblock In \emph{International Conference on Robotics and Automation (ICRA)}.
  {IEEE}, 2020.

\bibitem[Lozano{-}P{\'{e}}rez and Kaelbling(2014)]{lozanoPerez14constraint}
Tom{\'{a}}s Lozano{-}P{\'{e}}rez and Leslie~Pack Kaelbling.
\newblock A constraint-based method for solving sequential manipulation
  planning problems.
\newblock In \emph{Proc. of the Int. Conf. on Intelligent Robots and Systems
  (IROS)}, 2014.

\bibitem[Mason(1985)]{1985-Mason-mechanicsmanipulation}
Matthew Mason.
\newblock The mechanics of manipulation.
\newblock In \emph{Int.{} Conf.{} on Robotics and {{Automation}} (ICRA'85)}.
  {IEEE}, 1985.

\bibitem[Okada et~al.(2017)Okada, Rigazio, and Aoshima]{okada2017path}
Masashi Okada, Luca Rigazio, and Takenobu Aoshima.
\newblock Path integral networks: End-to-end differentiable optimal control.
\newblock \emph{arXiv preprint arXiv:1706.09597}, 2017.

\bibitem[Pascanu et~al.(2017)Pascanu, Li, Vinyals, Heess, Buesing,
  Racani{\`{e}}re, Reichert, Weber, Wierstra, and Battaglia]{pascanu17learning}
Razvan Pascanu, Yujia Li, Oriol Vinyals, Nicolas Heess, Lars Buesing,
  S{\'{e}}bastien Racani{\`{e}}re, David~P. Reichert, Theophane Weber, Daan
  Wierstra, and Peter~W. Battaglia.
\newblock Learning model-based planning from scratch.
\newblock \emph{CoRR}, abs/1707.06170, 2017.

\bibitem[Paxton et~al.(2019)Paxton, Barnoy, Katyal, Arora, and
  Hager]{paxton19visual}
Chris Paxton, Yotam Barnoy, Kapil~D. Katyal, Raman Arora, and Gregory~D. Hager.
\newblock Visual robot task planning.
\newblock In \emph{International Conference on Robotics and Automation (ICRA)},
  pages 8832--8838. {IEEE}, 2019.

\bibitem[Racani{\`e}re et~al.(2017)Racani{\`e}re, Weber, Reichert, Buesing,
  Guez, Rezende, Badia, Vinyals, Heess, Li, et~al.]{racaniere2017imagination}
S{\'e}bastien Racani{\`e}re, Th{\'e}ophane Weber, David Reichert, Lars Buesing,
  Arthur Guez, Danilo~Jimenez Rezende, Adria~Puigdom{\`e}nech Badia, Oriol
  Vinyals, Nicolas Heess, Yujia Li, et~al.
\newblock Imagination-augmented agents for deep reinforcement learning.
\newblock In \emph{Advances in neural information processing systems}, 2017.

\bibitem[Rodriguez et~al.(2019)Rodriguez, Nottensteiner, Leidner, Kasecker,
  Stulp, and Albu{-}Sch{\"{a}}ffer]{rodriguez19iteratively}
Ismael Rodriguez, Korbinian Nottensteiner, Daniel Leidner, Michael Kasecker,
  Freek Stulp, and Alin Albu{-}Sch{\"{a}}ffer.
\newblock Iteratively refined feasibility checks in robotic assembly sequence
  planning.
\newblock \emph{Robotics and Automation Letters}, 2019.

\bibitem[Silver et~al.(2017)Silver, van Hasselt, Hessel, Schaul, Guez, Harley,
  Dulac-Arnold, Reichert, Rabinowitz, Barreto, et~al.]{silver2017predictron}
David Silver, Hado van Hasselt, Matteo Hessel, Tom Schaul, Arthur Guez, Tim
  Harley, Gabriel Dulac-Arnold, David Reichert, Neil Rabinowitz, Andre Barreto,
  et~al.
\newblock The predictron: End-to-end learning and planning.
\newblock In \emph{International Conference on Machine Learning}, 2017.

\bibitem[Srinivas et~al.(2018)Srinivas, Jabri, Abbeel, Levine, and
  Finn]{srinivas18universal}
Aravind Srinivas, Allan Jabri, Pieter Abbeel, Sergey Levine, and Chelsea Finn.
\newblock Universal planning networks: Learning generalizable representations
  for visuomotor control.
\newblock In \emph{International Conference on Machine Learning (ICML)}, pages
  4739--4748, 2018.
\newblock URL \url{http://proceedings.mlr.press/v80/srinivas18b.html}.

\bibitem[Srivastava et~al.(2014)Srivastava, Fang, Riano, Chitnis, Russell, and
  Abbeel]{srivastava14combined}
Siddharth Srivastava, Eugene Fang, Lorenzo Riano, Rohan Chitnis, Stuart~J.
  Russell, and Pieter Abbeel.
\newblock Combined task and motion planning through an extensible
  planner-independent interface layer.
\newblock In \emph{Proc. of the Int. Conf. on Robotics and Automation (ICRA)},
  2014.

\bibitem[Tamar et~al.(2016)Tamar, Wu, Thomas, Levine, and
  Abbeel]{tamar2016value}
Aviv Tamar, Yi~Wu, Garrett Thomas, Sergey Levine, and Pieter Abbeel.
\newblock Value iteration networks.
\newblock In \emph{Advances in Neural Information Processing Systems}, pages
  2154--2162, 2016.

\bibitem[Toussaint(2015)]{toussaint15lgp}
Marc Toussaint.
\newblock Logic-geometric programming: An optimization-based approach to
  combined task and motion planning.
\newblock In \emph{Proceedings of the Twenty-Fourth International Joint
  Conference on Artificial Intelligence, {IJCAI}}, pages 1930--1936. {AAAI}
  Press, 2015.
\newblock URL \url{http://ijcai.org/Abstract/15/274}.

\bibitem[Toussaint and Lopes(2017)]{toussaint17mbts}
Marc Toussaint and Manuel Lopes.
\newblock Multi-bound tree search for logic-geometric programming in
  cooperative manipulation domains.
\newblock In \emph{International Conference on Robotics and Automation (ICRA)},
  pages 4044--4051. {IEEE}, 2017.

\bibitem[Toussaint et~al.(2018)Toussaint, Allen, Smith, and
  Tenenbaum]{18-toussaint-RSS}
Marc Toussaint, Kelsey~R Allen, Kevin~A Smith, and Josh~B Tenenbaum.
\newblock Differentiable physics and stable modes for tool-use and manipulation
  planning.
\newblock In \emph{Proc{.} of Robotics: Science and Systems (R:SS)}, 2018.

\bibitem[Toussaint et~al.(2020)Toussaint, Ha, and
  Driess]{20-toussaint-physicsLGP}
Marc Toussaint, Jung-Su Ha, and Danny Driess.
\newblock Describing physics for physical reasoning: Force-based sequential
  manipulation planning.
\newblock arXiv:2002.12780, 2020.

\bibitem[Wang et~al.(2018)Wang, Garrett, Kaelbling, and
  Lozano-P{\'e}rez]{wang2018active}
Zi~Wang, Caelan~Reed Garrett, Leslie~Pack Kaelbling, and Tom{\'a}s
  Lozano-P{\'e}rez.
\newblock Active model learning and diverse action sampling for task and motion
  planning.
\newblock In \emph{International Conference on Intelligent Robots and Systems
  (IROS)}, pages 4107--4114. IEEE, 2018.

\bibitem[Watter et~al.(2015)Watter, Springenberg, Boedecker, and
  Riedmiller]{watter15embed}
Manuel Watter, Jost~Tobias Springenberg, Joschka Boedecker, and Martin~A.
  Riedmiller.
\newblock Embed to control: {A} locally linear latent dynamics model for
  control from raw images.
\newblock In \emph{Advances in Neural Information Processing Systems}, pages
  2746--2754, 2015.

\bibitem[Wells et~al.(2019)Wells, Dantam, Shrivastava, and
  Kavraki]{wells2019learning}
Andrew~M Wells, Neil~T Dantam, Anshumali Shrivastava, and Lydia~E Kavraki.
\newblock Learning feasibility for task and motion planning in tabletop
  environments.
\newblock \emph{Robotics and Automation Letters}, 4\penalty0 (2):\penalty0
  1255--1262, 2019.

\bibitem[Wilson and Hermans(2019)]{wilson2019collections}
Matthew Wilson and Tucker Hermans.
\newblock Learning to manipulate object collections using grounded state
  representations.
\newblock \emph{Conference on Robot Learning}, 2019.
\newblock URL \url{https://arxiv.org/abs/1909.07876}.

\bibitem[Xie et~al.(2019)Xie, Ebert, Levine, and Finn]{xieimprovisation}
Annie Xie, Frederik Ebert, Sergey Levine, and Chelsea Finn.
\newblock Improvisation through physical understanding: Using novel objects as
  tools with visual foresight.
\newblock In \emph{Proc{.} of Robotics: Science and Systems (R:SS)}, 2019.

\end{thebibliography}

\end{document}